\documentclass[10pt]{article}
\usepackage{spconf}
\usepackage{amsmath,graphicx}
\usepackage{amssymb,comment,amsthm,amsfonts}
\usepackage{algorithm,algcompatible}
\usepackage{subfig}
\usepackage{caption}
\usepackage{cite}
\usepackage{url}
\usepackage{algorithm}
\algblockdefx{FORALLP}{ENDFAP}[1]%
{\textbf{for all }#1 \textbf{do in parallel}}%
{\textbf{end for}}

\newcommand{\maximize}{\mathop{\rm maximize}\limits}
\newcommand{\minimize}{\mathop{\rm minimize}\limits}
\newcommand{\argmin}{\mathop{\rm arg~min}\limits}

\usepackage{xcolor}
\title{Dynamic  Scheduling for Federated Edge Learning with Streaming Data}
\name{Chung-Hsuan Hu, Zheng Chen, and Erik G. Larsson \thanks{This work was supported in part by Zenith, ELLIIT, and the Knut and Alice Wallenberg (KAW) Foundation.}}
\address{Department of Electrical Engineering (ISY), Linköping University, 581 83 Linköping, Sweden}
\begin{document}\ninept
	%
	\maketitle
	\begin{abstract}
		In this work, we consider a Federated Edge Learning (FEEL) system where training data are randomly generated over time at a set of distributed edge devices with long-term energy constraints. Due to limited communication resources and latency requirements, only a subset of devices is scheduled for participating in the local training process in every iteration.
		We formulate a stochastic network optimization problem for designing a dynamic scheduling policy that maximizes the time-average data importance from scheduled user sets subject to energy consumption and latency constraints. Our proposed algorithm based on the Lyapunov optimization framework outperforms alternative methods without considering time-varying data importance, especially when the generation of training data shows strong temporal correlation. 
	\end{abstract}
	\section{Introduction}
	Federated learning (FL) over wireless networks is an emerging research direction that lies within the intersection between wireless communications and machine learning. Particularly, in Federated Edge Learning (FEEL) where a large set of wireless edge devices participate in a common model training task, the limitation of wireless communication resources (e.g., frequency, time, energy) can greatly affect the efficiency of model aggregation and the learning performance. The heterogeneity of devices in terms of training data distribution, channel condition, and computing capability makes the optimal scheduling and resource allocation design a challenging task. 
	
	In the literature of device scheduling and resource allocation for FEEL systems, most existing work focuses on the heterogeneity of data and/or wireless channels, without consideration of heterogeneous computation capability and energy availability \cite{rizk2022fed,amiriconvergence,wu2022node,hu2023sch,salehi2021fl,Malandrino2021fl,zhang2022comm}.
	In a FEEL system, the local process consists of two phases: model training and update transmission. Devices with superior computing capability can finish their model training faster; however, such computing strength, in terms of CPU (Central Processing Unit) cycle frequency, might be time-varying depending on the concurrent activities of a device. Adopting a higher operating CPU frequency leads to more energy consumption, which is crucial for battery-limited devices. 
	On the other hand, energy consumption in the model transmission phase is affected by power control and transmission time constraint. Therefore, an optimal scheduling design should take into account all the system dynamics in channel condition, computing capability, and energy consumption. Some existing works have studied device scheduling in FEEL with a limited energy budget \cite{energy-aware,sun2021dynamic,xu2021client}, or further with constrained latency \cite{guo2022dynamicSch,chen2020ajoint}, while others focus on minimizing energy consumption \cite{li2021opt, alba2021findGrain}, or optimizing both energy and time efficiency \cite{luo2021costeffective,wan2021ca}, for a given accuracy level of learning performance.
	Moreover, \cite{zheng2021fed,ji2022client,yu2022jointly} adopt alternative approaches, which jointly consider learning and system efficiency from energy and/or time aspects as their objectives in the optimization. 
	
	However, all these existing methods focus on the static training data scenario, which means that all the local data are available at the beginning of the training process. 
	In practical scenarios, the training data might be generated randomly over time, which leads to time-varying local loss functions  \cite{chen2020asynchronous,damaskinos2020fleet,mitra2021ofl}.
	To avoid over-fitting in model training, the statistics of the newly arrived data with respect to the entire data collection process should be considered in the scheduling design.
	
	The main novelty of this work is that we consider a FEEL system with streaming data generation at wireless edge devices. We formulate a stochastic network optimization problem and proposes a dynamic scheduling algorithm that jointly considers the data importance, per-round latency requirements, and time-average energy constraints. Similar stochastic optimization approaches have been adopted in \cite{energy-aware,sun2021dynamic,guo2022dynamicSch,ji2022client,xu2021client}. The objective function is designed based on an importance-aware metric that ensures robust learning performance under heterogeneous data distributions and their arrival patterns across different devices. The effectiveness of the proposed design is validated by numerical simulations.
	
	\section{System Model}
	We consider a FEEL system with $K$ edge devices participating in training a global model $\boldsymbol{\theta}\in\mathbb{R}^d$. We denote the device set by $\mathcal{K}=\{1,...,K\}$, where each device $k\in\mathcal{K}$ has a local training data set $\mathcal{S}_k$. The objective of the system is to minimize an empirical loss function
	\begin{equation}
		F(\boldsymbol{\theta})=\sum_{k\in\mathcal{K}}\frac{|\mathcal{S}_k|}{|\cup_{j\in\mathcal{K}}\mathcal{S}_j|} F_k(\boldsymbol{\theta}),
		\label{eq:globalLoss_f}
	\end{equation}
	where $F_k(\boldsymbol{\theta})$ is the local loss function at device $k$.
	New training samples are generated randomly over time following some stochastic processes. At time instant $t$, only $\mathcal{S}_k(t)=\mathcal{S}_k(t-1)\cup\mathcal{B}_k(t)$ is available for local training, where $\mathcal{B}_k(t)$ is the newly arrived data set after the previous time instant, $\mathcal{S}_k(0)=\emptyset$ and $\lim\limits_{t\rightarrow\infty}\mathcal{S}_k(t)=\mathcal{S}_k$.\footnote{
	We may also consider another setting with $S_k(t)=\mathcal{S}_k(t-1)\cup\mathcal{B}_k(t)\setminus\mathcal{D}_k(t)$, where $D_k(t)$ denotes the set of deleted data in every time instance t.}
	Therefore, we minimize a time-varying loss function
	\begin{equation}
		F(\boldsymbol{\theta},t)=\sum_{k\in\mathcal{K}}\frac{|\mathcal{S}_k(t)|}{|\cup_{j\in\mathcal{K}}\mathcal{S}_j(t)|} F_k(\boldsymbol{\theta},t),
		\label{eq:globalLoss_t}
	\end{equation}
	where $F_k(\boldsymbol{\theta},t)$ is evaluated based on $\mathcal{S}_k(t)$, $\lim\limits_{t\rightarrow\infty}F_k(\boldsymbol{\theta},t)=F_k(\boldsymbol{\theta})$, and $\lim\limits_{t\rightarrow\infty}F(\boldsymbol{\theta},t)=F(\boldsymbol{\theta})$.
	
	The training process consists of multiple communication rounds.
	In the $t$-th  round with $t=1,2,\ldots$, the following steps are executed:
	\begin{enumerate}
		\item The server broadcasts the current global model $\boldsymbol{\theta}(t)$ to the set of participating devices, which is denoted by $\Pi(t)$.
		\item Each device $k\in\Pi(t)$ runs a fixed number of mini-batch stochastic gradient descent (SGD) to obtain the model update $\boldsymbol{\triangle}\boldsymbol{\theta}_k(t)$, which is transmitted to the server.
		\item The server aggregates the received information and updates the global model
		\begin{equation}
			\boldsymbol{\theta}(t+1)=\sum_{k\in\Pi(t)}\frac{|\mathcal{S}_k(t)|}{|\cup_{j\in\Pi(t)}\mathcal{S}_j(t)|}\boldsymbol{\triangle}\boldsymbol{\theta}_k(t)+\boldsymbol{\theta}(t).
			\label{eq:syncFlAggregation}	
		\end{equation}
	\end{enumerate}
	\subsection{Energy Consumption Model}
	The energy consumption of the $k$-th device in the $t$-th communication round can be written as
	\begin{equation}
		E_k(t)=E_k^{\text{cmp}}(t)+E_k^{\text{tr}}(t),
		\label{eq:Ekt}
	\end{equation}
	where $E_k^{\text{cmp}}(t)$ is the energy consumed from computation and $E_k^{\text{tr}}(t)$ is the energy consumption for transmission.
	
	\subsubsection{Energy for Local Computation}
	We apply dynamic voltage and frequency scaling (DVFS) to adjust the effectively used computation resource of a CPU.
	Let $f_k(t)$ represent the CPU clock frequency of the $k$-th device in the $t$-th round. 
	The energy consumption for the model update computation is approximately given by \cite{larsson2011impact}
	\begin{equation}
		E_k^{\text{cmp}}(t)=\lambda cf_k^2(t),
		\label{eq:Ec}
	\end{equation}
	where $\lambda$ is a power coefficient and $c$ is the required number of CPU cycles for computing a fixed number of mini-batch SGD.
	
	\subsubsection{Energy for Update Transmission}
	To transmit the model updates to the server, the entire bandwidth $B$ is shared among the participating devices. We define $\rho_k(t)$ as the bandwidth fraction assigned to the $k$-th device in the $t$-th round, where $\sum_{k\in\Pi(t)}\rho_k(t)=1$. Also, $P_k(t)$ is the transmit power. The achievable rate is
	\begin{equation}
		R_k(t)=\rho_k(t)B\log_2 \left(1+\frac{P_k(t)|g_k(t)|^2}{\rho_k(t)BN_0}\right),
		\label{eq:R}
	\end{equation}
	where $g_k(t)$ is the channel gain with $\mathbb{E}[|g_k(t)|^2]=\beta_k$ and $N_0$ is the spectral density of the noise.
	Assuming that the model updates are compressed and quantized with $S$ bits, the required transmission time is 
	\begin{equation}
		T^{\text{tr}}_k(t)=\frac{S}{R_k(t)}.
		\label{eq:Tt}
	\end{equation}
	The energy consumption for transmission is thus
	\begin{equation}
		E^{\text{tr}}_k(t)=P_k(t)T^{\text{tr}}_k(t).
		\label{eq:Et}
	\end{equation}
	
	\subsection{Latency Model}
	Define $T^{\text{cmp}}_k( t)$ as the time for model update computation; we have
	\begin{equation}
		T^{\text{cmp}}_k( t)=\frac{c}{f_k(t)}.
		\label{eq:Tc}
	\end{equation}
	Based on \eqref{eq:Tt} and \eqref{eq:Tc}, the latency of device $k$ to complete transmission and computation in the $t$-th round is
	\begin{equation*}
		T_k(t) = T^{\text{cmp}}_k(t)+T^{\text{tr}}_k(t).
	\end{equation*}

	\section{Problem Formulation}
	As we consider a streaming data setup, at any iteration $t$ the accumulated training data $\mathcal{S}_k(t)$ can be highly heterogeneous over time, even though its asymptotic counterpart, $\mathcal{S}_k$, is homogeneous across different devices. 
	To schedule devices with the highest impact on the learning performance, a natural choice is to prioritize those with lower similarity to the existing data and higher amount of newly arrived data since the last model pull. Thus, we define a data importance metric $I_k(t)$ for each device $k$,
	\begin{equation}
		\label{eq:iptc}
		I_k(t)=\frac{|\Pi_f(t)|\cdot|\mathcal{B}_k(t)|}{\sum_{j\in\Pi_f(t)}|\mathcal{B}_j(t)|}+\boldsymbol{1}\{t>1\}\cdot\frac{||\boldsymbol{x}-\boldsymbol{y}_k||_2^2}{||\boldsymbol{x}||_2^2+||\boldsymbol{y}_k||_2^2}.
	\end{equation}
	The first term in \eqref{eq:iptc} quantifies the proportion of newly arrived data among a candidate device set $\Pi_f(t)\subseteq\mathcal{K}$\footnote{The definition of $\Pi_f(t)$ will be introduced in Sec. \ref{Sec:algo}}, and the other examines the feature dissimilarity between the data that have been utilized, $\boldsymbol{x}$, and those newly generated, $\boldsymbol{y}_k$, by computing the normalized Euclidean distance between them.\footnote{Let $L(\mathcal{A})=[l_1,...,l_m]$ be an $m$-size feature vector and $\bar{L}(\mathcal{A})=(\sum_{i=1}^{m}l_i)/m$ be its average. Then, we define $\boldsymbol{x}=\left[L(\mathcal{X})-\bar{L}(\mathcal{X})\right]/\bar{L}(\mathcal{X})$ and $\boldsymbol{y}_k=\left[L(\mathcal{B}_k(t))-\bar{L}(\mathcal{B}_k(t))\right]/\bar{L}(\mathcal{B}_k(t))$ respectively, where $\mathcal{X}=\cup_{k\in\mathcal{K}}\mathcal{S}_k(\hat{t}_k)$ and $\hat{t}_k=\max\{\tau|\tau\leq t-1,k\in\Pi(\tau)\}$.}

	To accelerate the FL process, and in the meantime maintain energy efficiency, we formulate a stochastic optimization problem as follows,
	\begin{subequations}
		\label{eq:opt} 
		\begin{align}	
			\underset{\Pi(t)}{\maximize}~~& \limsup\limits_{T\rightarrow \infty}\frac{1}{T} \sum_{t=1}^{T}\mathbb{E}\left[\sum_{k\in \Pi(t)}  I_k(t)\right],\\
			\textrm{subject~to}~~& \limsup\limits_{T\rightarrow \infty}\frac{1}{T} \sum_{t=1}^{T}\mathbb{E}\left[E_k(t)\right]\leq E^{\text{avg}}_{k}, \label{eq:const1}\\
			& T_k(t)\leq T_{\text{rd}},\forall k\in\Pi(t), \label{eq:const2}\\
			& \Pi(t)\subseteq \mathcal{K}.\label{eq:const4}  			
		\end{align}
	\end{subequations}
	Here, the expectation is subject to the randomness in data generation, wireless link quality, and computing power availability.
	For any device $k$, the constraint \eqref{eq:const1} reflects the long-term energy constraint; the latency constraint \eqref{eq:const2} ensures that every global round can be finished within a certain time window. 
	We assume equal bandwidth allocation among participating devices. With the knowledge of $\mathbb{E}[|g_k(t)|^2]=\beta_k$, we adopt the transmit power as $P_k(t)=P_0/\beta_k$ for some constant $P_0>0$ such that the expected signal-to-noise ratio remains the same for all the devices.
	
	\section{Dynamic User Scheduling Algorithm}
	\label{Sec:algo}
	We use Lyapunov optimization framework to solve the stochastic network optimization problem presented in \eqref{eq:opt} \cite{neely2010stochastic}. We define a virtual queue $Q_k(t)$ for the constraint \eqref{eq:const1}, which evolves as
	\begin{equation}
		Q_k(t+1)=\max\left[Q_k(t)+s_k(t)E_k(t)-E_k^{\text{avg}},0\right],
		\label{eq:virtualQ}
	\end{equation} 
	where $s_k(t)=\boldsymbol{1}\{k\in\Pi(t)\}$.
	Then, \eqref{eq:opt} can be transformed into a queue stability problem\footnote{The derivation will be included in an extended version of this paper.}
	\begin{align}
		\minimize~~&\sum_{k=1}^KQ_k(t)s_k(t)E_k(t)-V\sum_{k=1}^Ks_k(t)I_k(t),\label{obj:minDPP}\\
		\textrm{subject~to}~~&\eqref{eq:const2},\eqref{eq:const4}.\nonumber
	\end{align}
	Here, $V>0$ is a constant that balances the tradeoff between the optimization of queue stability and learning performance.
	We summarize all the steps in Algorithm \ref{alg:algo1} and give the details in the following subsections.
	\begin{algorithm}[t!]
		\caption{Learning-aware dynamic resource management}
		\label{alg:algo1}
		\begin{algorithmic}[1]
			\STATE Obtain $\mathcal{S}_k(t)$, $\beta_k$, $f_k(t)$, $Q_k(t)$, $\forall k\in\mathcal{K}$, $\gamma$, $T_{\text{rd}}$, $P_0$, $S$, $B$, $V$, $\lambda$, $c$, $N_0$, $|\Pi(t)|$, $\boldsymbol{\theta}(t)$;
			\STATE Find a feasible device set $\Pi_f(t)$ according to \eqref{ieq:ltc_const} and compute $I_k(t),\forall k\in\Pi_f(t)$. 
			\STATE Determine scheduling policy $\Pi^*(t)$ by solving \eqref{opt:scheduling}.
			\STATE Broadcast $\boldsymbol{\theta}(t)$ to $\Pi^*(t)$.
			\FORALLP{device $k\in\Pi^*(t)$}
			\STATE Fixed steps of mini-batch SGD.
			\ENDFAP
			\WHILE{all devices in $\Pi^*(t)$ complete local training}
			\STATE Acquire $g_k(t)$ and $E^{\text{cmp}}_k(t)$, $\forall k\in\Pi^*(t)$. 
			\STATE Obtain a feasible device subset  $\bar{\Pi}(t)\subseteq\Pi^*(t)$ by the approach described in Section \ref{prob:af_training}. 
			\STATE \textbf{break}
			\ENDWHILE
			\STATE Compute $E_k(t)=\begin{cases}E_k^{\text{cmp}}(t), & k\in\Pi^*(t)\setminus\bar{\Pi}(t)\\
				E_k^{\text{cmp}}(t)+E_k^{\text{tr}}(t), &k\in\bar{\Pi}(t) \end{cases}$
			\STATE $Q_k(t+1)\leftarrow Q_k(t)-E_k^{\text{avg}}+\boldsymbol{1}\{k\in\Pi^*(t)\}\cdot E_k(t)$
			\FORALLP{device $k\in\bar{\Pi}(t)$}
			\STATE Transmit $\boldsymbol{\triangle}\boldsymbol{\theta}_k(t)$ to the server with $\rho_k(t)=1/|\bar{\Pi}(t)|$ and $P_k(t)=P_0/\beta_k$.
			\ENDFAP
			\STATE Update $\boldsymbol{\theta}(t+1)$ according to \eqref{eq:syncFlAggregation} with $\Pi(t)=\bar{\Pi}(t)$
		\end{algorithmic}
	\end{algorithm}
	\subsection{Scheduling Phase: Determine $\Pi^*(t)$}
	\label{prob:bf_scheduling}
	Since all the devices in the scheduled set need to satisfy \eqref{eq:const2} and the channel gain $g_k(t)$ is unknown at this phase, we define a surrogate rate function
	\begin{equation}
		\tilde{R}_k(t)=\frac{\gamma B}{|\Pi(t)|}\log_2\left(1+\frac{P_0|\Pi(t)|}{BN_0}\right)
		\label{ieq:srgt_func}
	\end{equation}
	for computing the transmission time. In \eqref{ieq:srgt_func}, $\gamma\leq 1$ is introduced as a scaling factor inversely proportional to the time reserved for the future transmission.\footnote{Adopting a small $\gamma$ can be perceived as strategically underestimating the transmission rate while evaluating the latency constraint, which would reserve an extra time buffer for transmission in case the link quality varies before and after local training.}
	Then, we rearrange \eqref{obj:minDPP} as
	\begin{align}
		\underset{\Pi(t)}{\minimize}~~&\sum_{k\in\Pi(t)}\Big[Q_k(t)\lambda cf_k^2(t)-VI_k(t)\nonumber\\
		&+\frac{Q_k(t) S|\Pi(t)|P_0}{\gamma B\beta_k\log_2\left(1+\frac{|\Pi(t)|P_0}{BN_0}\right)}\Big],\label{opt:obj}\\
		\textrm{subject~to}~~& \Pi(t)\subseteq \mathcal{K},~\forall k\in\Pi(t),\nonumber\\
		&\frac{c}{f_k(t)}+\frac{S|\Pi(t)|}{\gamma B\log_2\left(1+\frac{|\Pi(t)|P_0}{BN_0}\right)}\leq T_{\text{rd}}.\label{ieq:ltc_const}
	\end{align}
	Denote by $\Pi_f(t)$ the feasible device subset that satisfies \eqref{ieq:ltc_const}. Then, we obtain the optimal scheduling policy 
	\begin{align}
		\Pi^*(t)=\underset{\Pi(t)\subseteq\Pi_f(t)}{\argmin}~~\text{\eqref{opt:obj}}\label{opt:scheduling}
	\end{align}
	\subsection{Aggregation Phase: Determine $\bar{\Pi}(t)$}
	\label{prob:af_training}
	When the local training of all the participating devices finishes and before the update transmission, we need to verify whether the remaining time $T_{\text{rd}}-T_k^{\text{cmp}}(t)$ for each scheduled device $k$ is sufficient for transmission, based on the knowledge of $g_k(t)$, $\forall k\in\Pi^*(t)$.
	We define a function $\mathcal{G}: \Pi\rightarrow\Pi^-$ that returns the infeasible device subset $\Pi^-\subseteq\Pi$ based on the latency constraint \eqref{eq:const2};
	\begin{equation*}
		\mathcal{G}(\Pi)=\Big\{k\big||g_k(t)|^2<\frac{3C_1\beta_k BN_0}{|\Pi|P_0},k\in\Pi\Big\},
	\end{equation*}
	where $C_1=2^{S|\Pi|/\left[B(T_{\text{rd}}-T_k^{\text{cmp}}(t))\right]}-1$. 
	Let $\bar{\Pi}(t)$ be initialized as $\Pi^*(t)$. If $|\mathcal{G}(\bar{\Pi}(t))|>0$, the device $j$ with the longest transmission time, i.e., $j=\argmin_{i\in\mathcal{G}(\bar{\Pi}(t))}|g_i(t)|^2/\beta_i$, is removed from $\bar{\Pi}(t)$.
	This step repeats until $\mathcal{G}(\bar{\Pi}(t))=\emptyset$.

	\section{Simulation Results}
	\label{sec:simulation_results}
	We simulate a FEEL system with $K=40$ devices using MNIST \cite{lecun-mnisthandwrittendigit-2010} as the local data to train a $d$-dimensional model $\boldsymbol{\theta}$ of a convolutional neural network, with $d=21840$, for solving a hand-written digit classification problem. 
	Details of the system setting are as follows.
	\begin{itemize}
		\item (Training data distribution) $60000$ data samples are distributed evenly to the devices, i.e., $|\mathcal{S}_k|=60000/K$. For the independent-and-identically-distributed (i.i.d.) case, the samples are randomly allocated to all the devices without replacement, while for the non-i.i.d. case, each device contains data with reduced digit variety.
		\item (Data arrival) Data arrive in the order of digit, and the first arriving digit is randomly picked at each device. The arrival timings follow truncated normal distribution with mean $\mu_k\sim\mathcal{U}(0,T_{\text{tot}})$ and a clipping range $[0,T_{\text{tot}}]$, where $T_{\text{tot}}$ denotes the entire execution time of the system.
		\item CPU frequency $f_k(t)\sim\mathcal{U}(0.02,1.52)$ GHz.
		\item Large-scale fading factor $\beta_k\sim\mathcal{U}(-5\text{dB},3\text{dB})$.
		\item Constant parameters are listed in Table \ref{tab:constParam}.
	\end{itemize}
	\begin{table}[h]
		\def\arraystretch{1.2}%
		\caption{System parameters} 
		\centering 
		\begin{tabular}{c rrrr} 
			\hline
			\textbf{Parameter}& \textbf{Value} &\textbf{Parameter} & \textbf{Value}\\ [0.3ex]
			\hline 
			eff. received power $P_0$ & $28$ dBm & bandwidth $B$ & $20$ MHz \\
			power coefficient $\lambda$ & $10^{-27}$ &	model size $S$ &  $32d$\\
			computation scaling $c$ & $600\cdot32d$ & noise power $N_0$ & $10^{-13}$ W\\
			latency bound $T_{\text{rd}}$ &$4$ sec \\
			avg. energy $E_k^{\text{avg}},\forall k$ &$0.0005$ J\\[1ex] 
			\hline 
		\end{tabular}
		\label{tab:constParam}
	\end{table}
	\begin{figure}[h]
		\centering
		\subfloat[i.i.d.]{\includegraphics[width=.5\columnwidth]{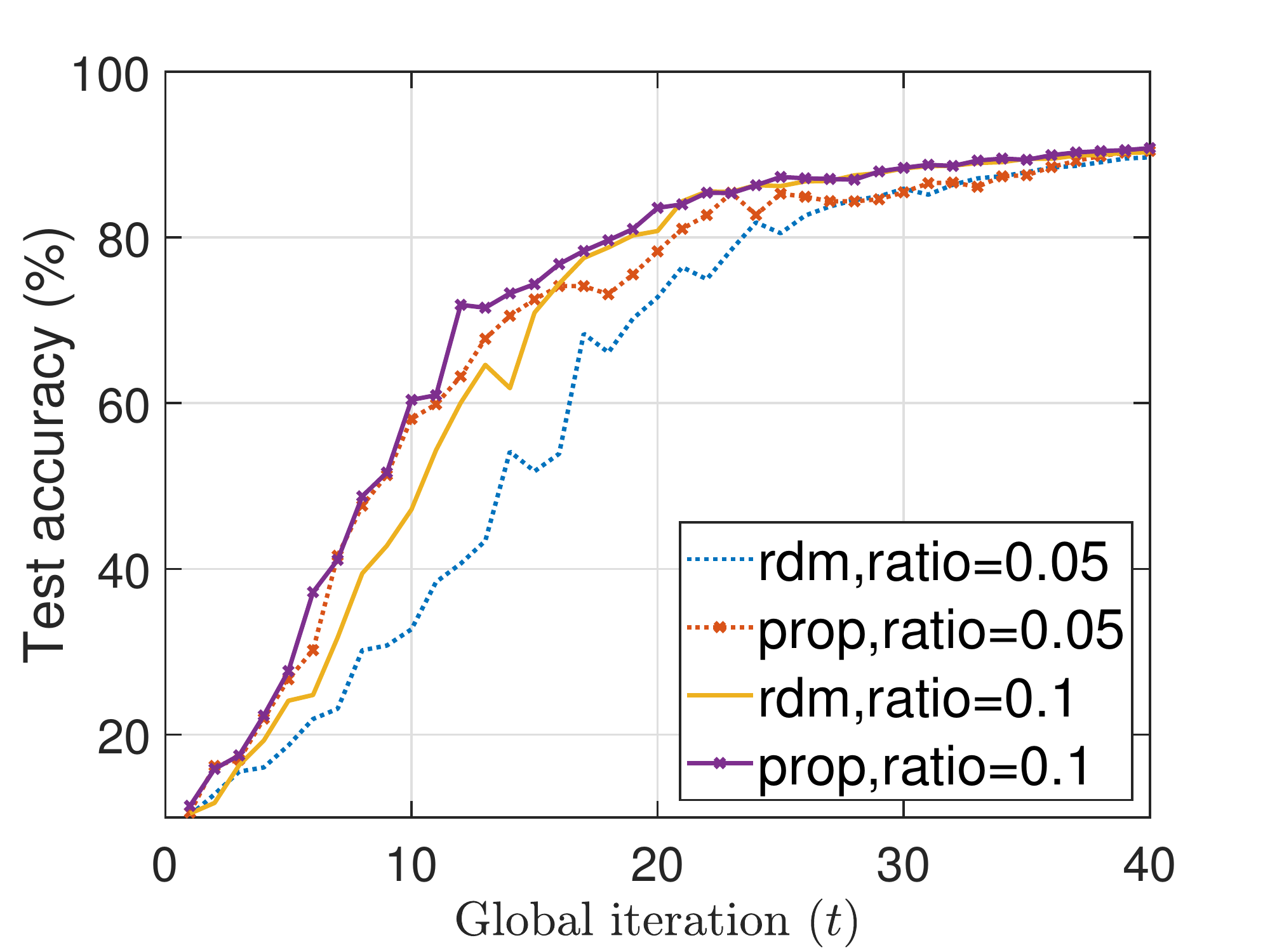}\label{fig:testAccu_iid}}
		\subfloat[non-i.i.d.]{\includegraphics[width=.5\columnwidth]{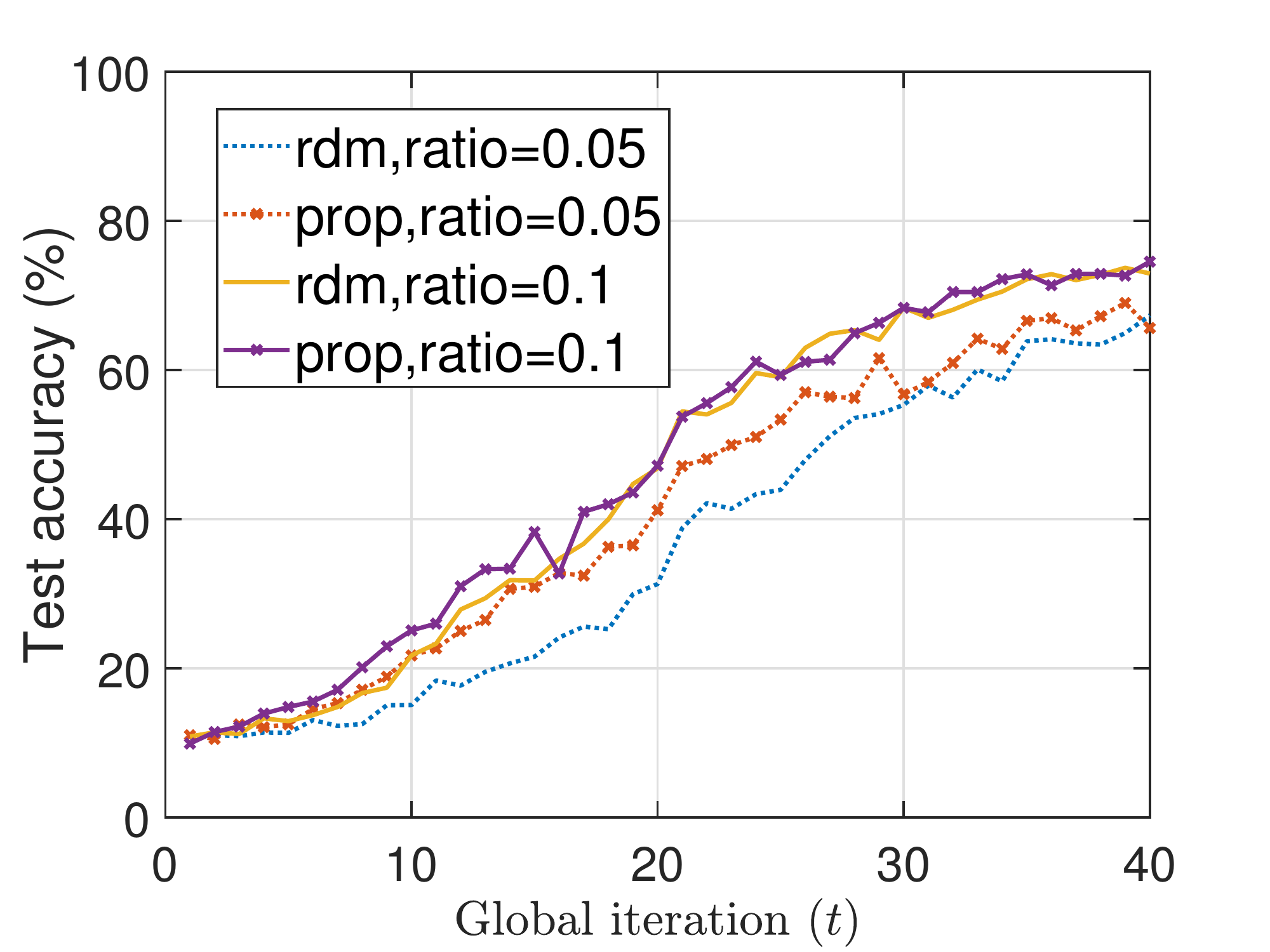}\label{fig:testAccu_niid}}\\
		\subfloat[i.i.d.]{\includegraphics[width=.5\columnwidth]{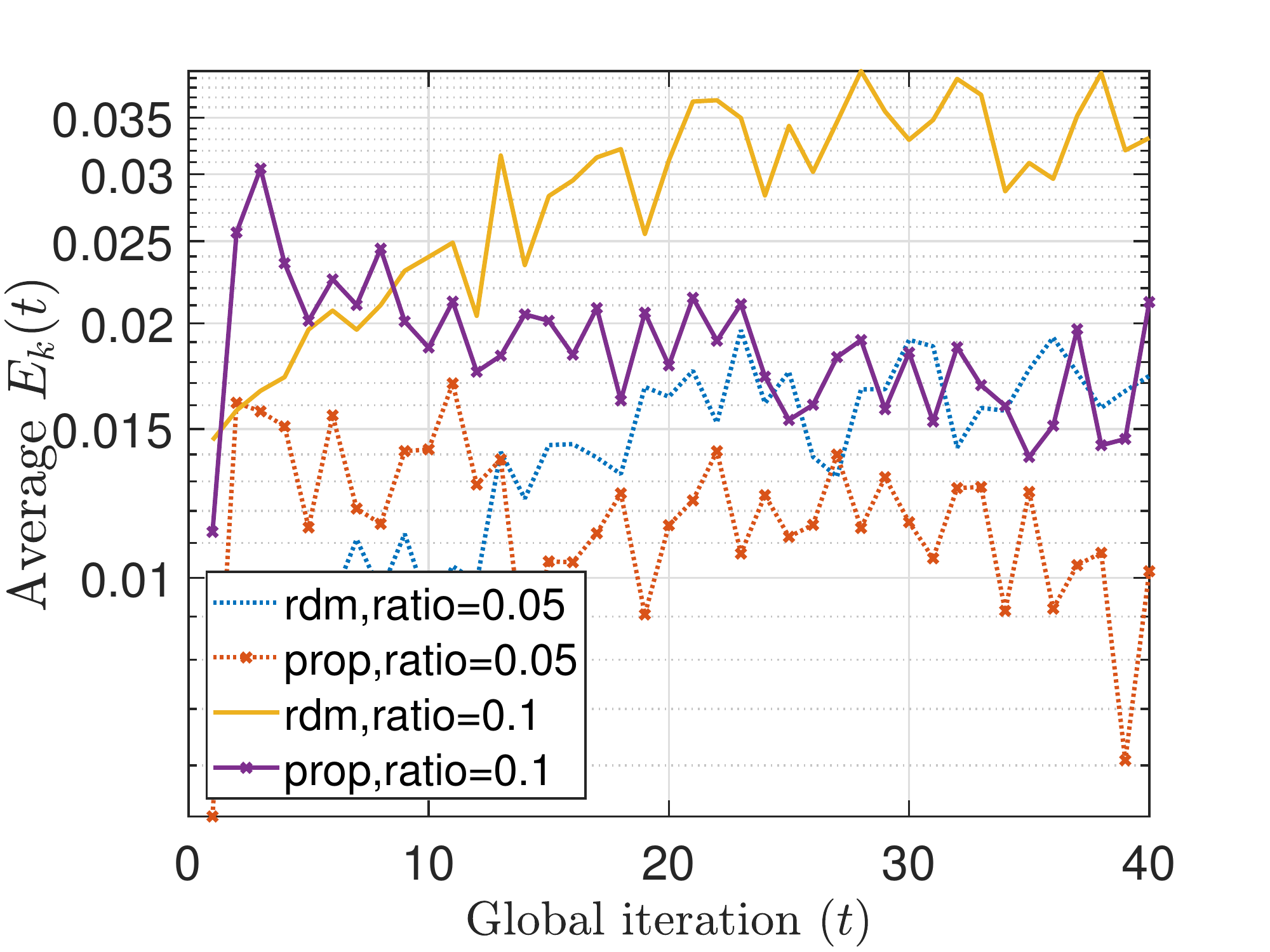}\label{fig:avgE_SameRatio_iid}}
		\subfloat[non-i.i.d.]{\includegraphics[width=.5\columnwidth]{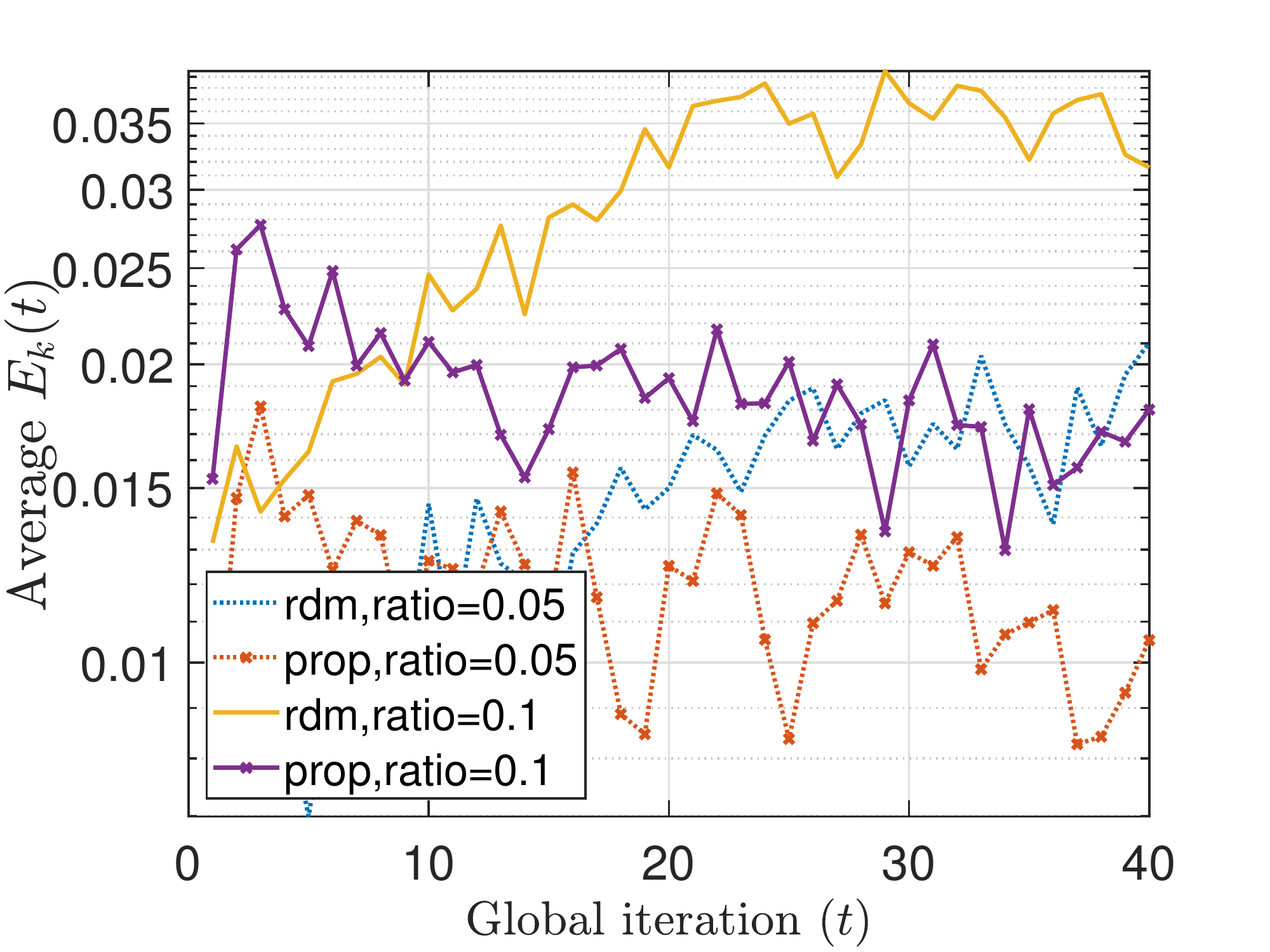}\label{fig:avgE_SameRatio_niid}}
		\caption{Test accuracy and energy consumption comparison between the proposed ('prop') and the random ('rdm') methods, with different scheduling ratios of $|\Pi(t)|/K$ and $V=0.05$. For the non-i.i.d. case, each device contains up to $3$ unique digits.} 
	\label{fig:testSameRatio}
	\hfill
\end{figure}
\subsection{Improvement in Learning and Energy Efficiency}
We compare the performance of our proposed design with a baseline method that adopts random scheduling of devices that satisfies the per-round latency requirement. 
The comparisons of test accuracy are shown in Figs. \ref{fig:testAccu_iid} and \ref{fig:testAccu_niid}, under both i.i.d. and non-i.i.d. data settings.
As observed from the simulation results, our method has better test accuracy in the i.i.d. scenario because the proposed scheduling policy prioritizes the devices with higher amounts of fresher data. It also outperforms the alternative method in the non-i.i.d. scenario as the variety of new data is considered in the scheduling criteria, which avoids over-fitting problem by reducing bias in the aggregated model.
Moreover, in Figs. \ref{fig:avgE_SameRatio_iid} and \ref{fig:avgE_SameRatio_niid}, the comparisons of average per-device energy consumption show that our method consumes less over the learning process. Specifically, we observe $16\%$ and $35\%$ of reduction in power consumption for the case with scheduling ratios $|\Pi(t)|/K=0.05$ and $|\Pi(t)|/K=0.1$, respectively.
\subsection{Effectiveness of the Data Importance Metric}
\begin{figure}[t!]
	\centering
	\subfloat[i.i.d.]{\includegraphics[width=.5\columnwidth]{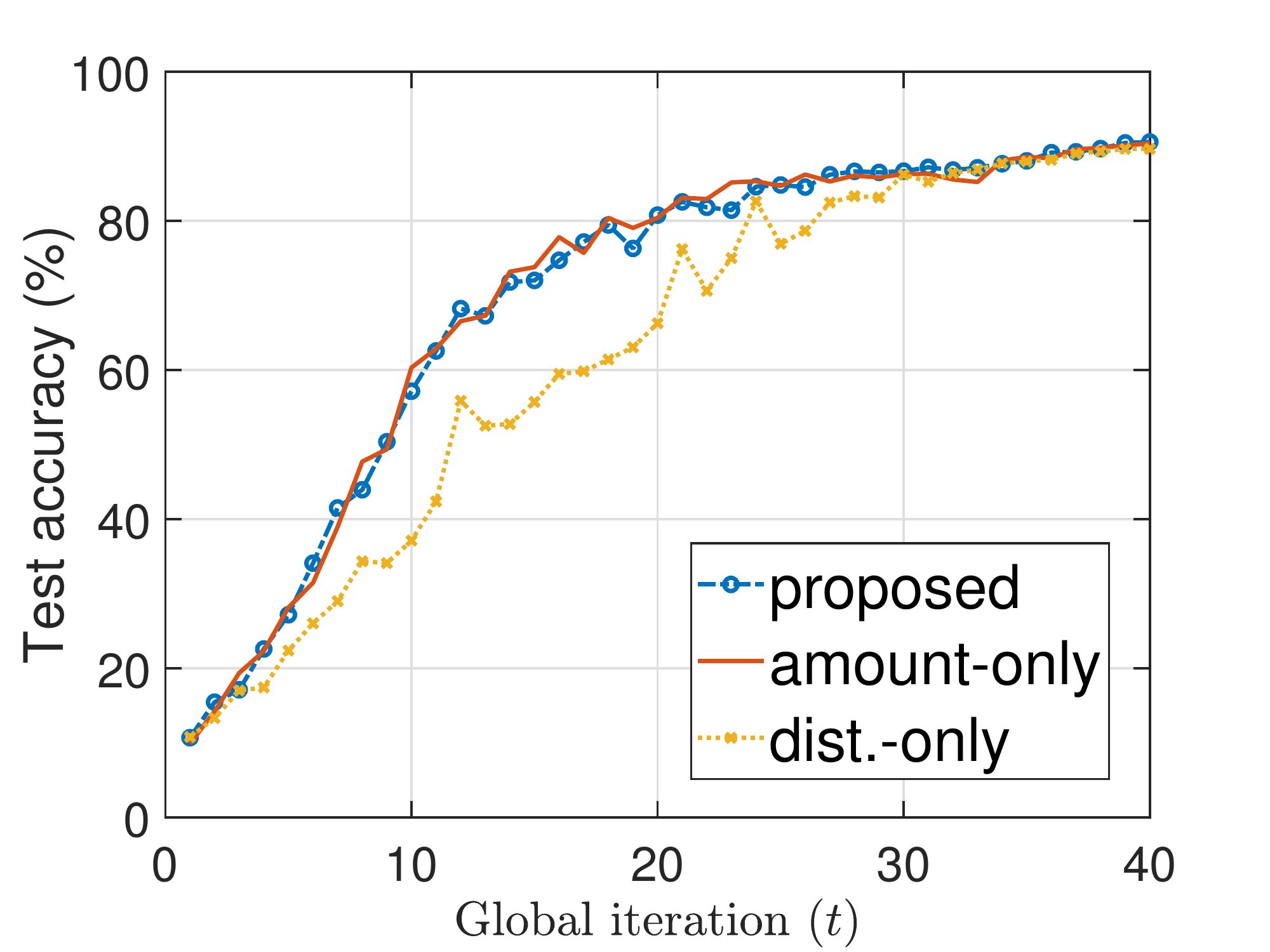}\label{fig:testAccu_m_iid}}
	\subfloat[non-i.i.d.]{\includegraphics[width=.5\columnwidth]{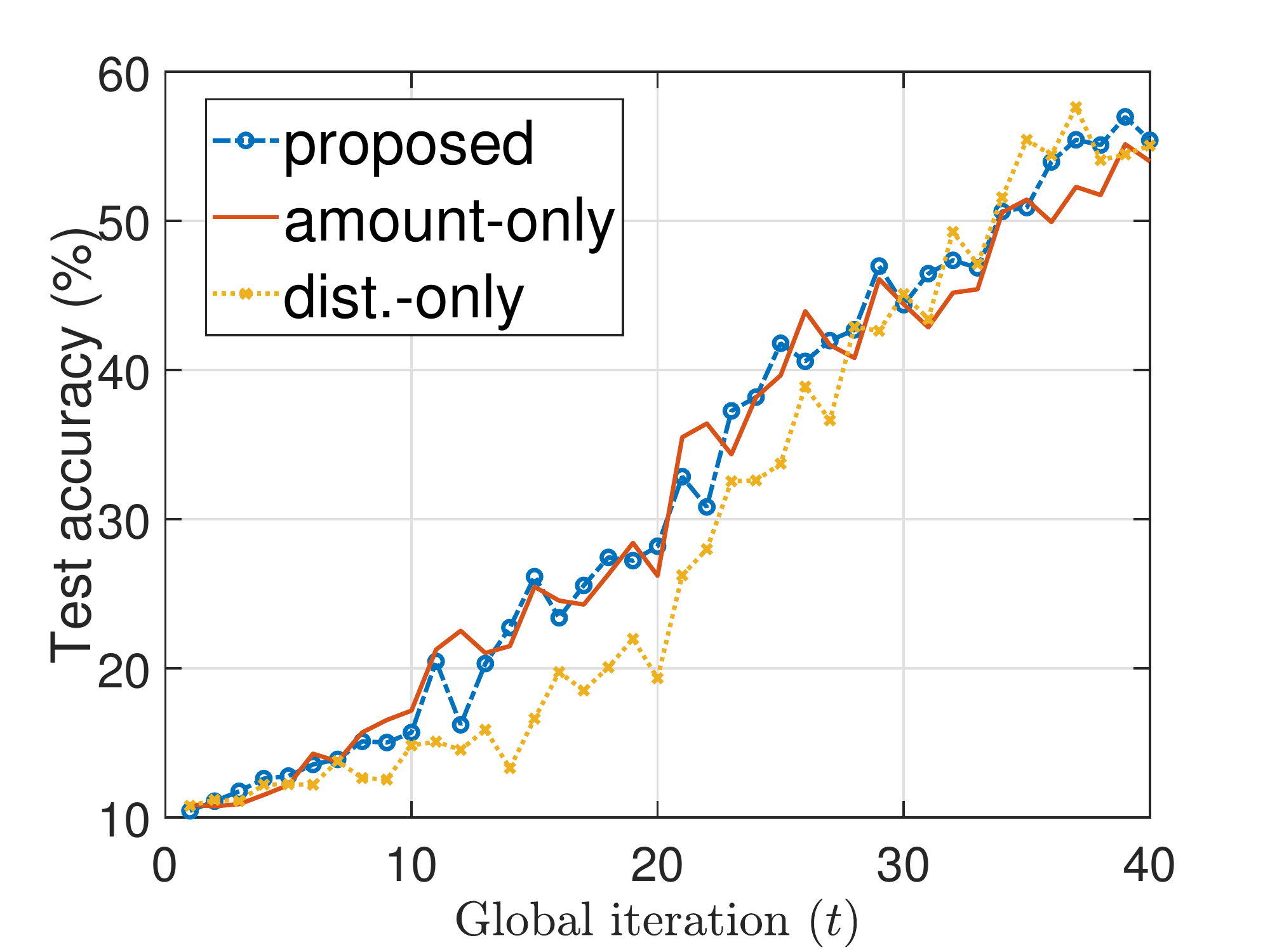}\label{fig:testAccu_m_niid}}
	\caption{Test accuracy based on different data importance metrics. $|\Pi(t)|/K=0.05$. For the non-i.i.d. case, each device contains up to $2$ unique digits.}
	\label{fig:metricCmpDist8}
	\hfill
\end{figure}

\begin{figure}[t!]
	\centering
	\subfloat[truncated normal distribution ]{\includegraphics[width=.5\columnwidth]{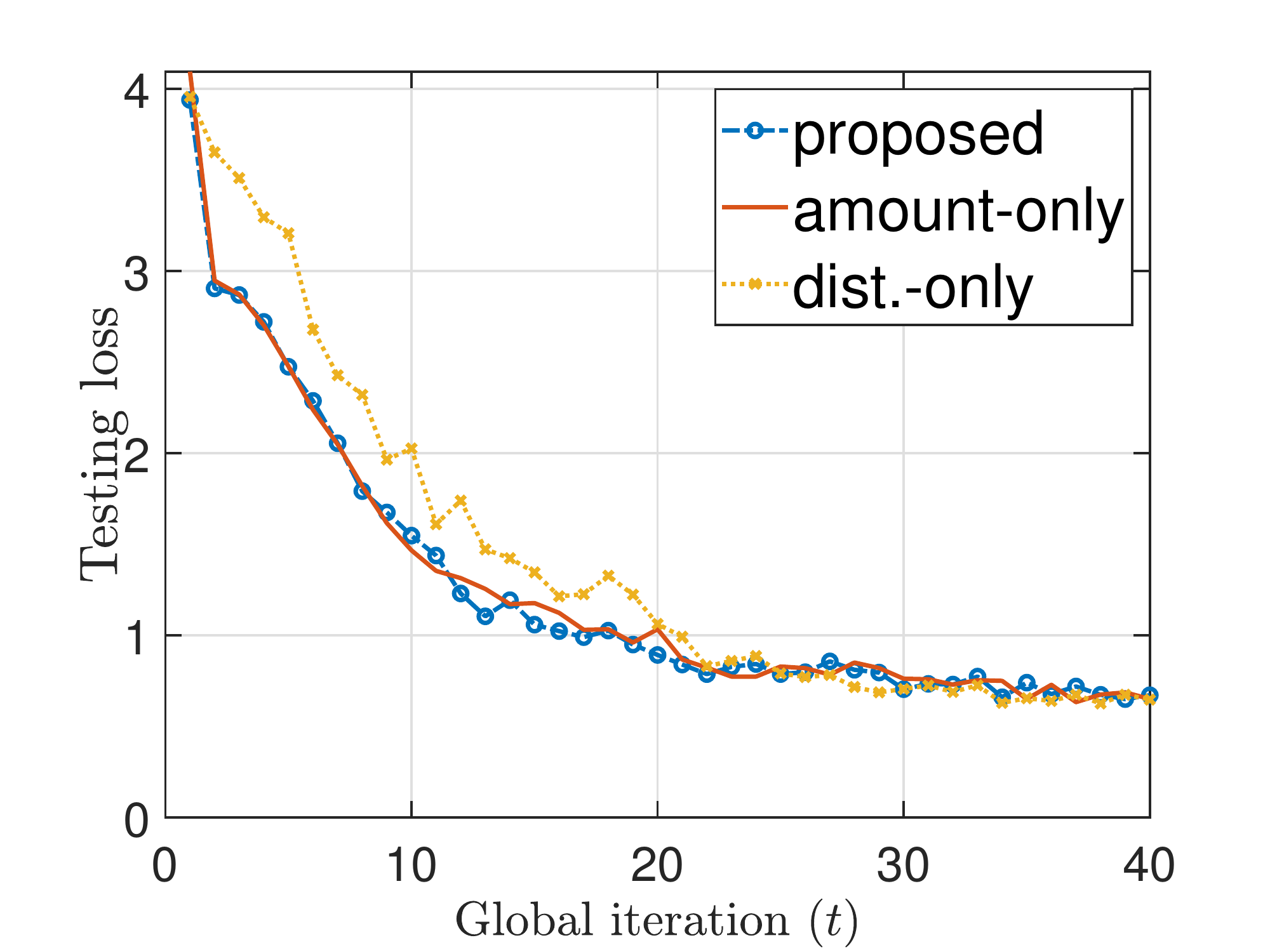}\label{fig:testLoss_dist8_niid}}
	\subfloat[uniform distribution in $(0,T_{\text{tot}})$]{\includegraphics[width=.5\columnwidth]{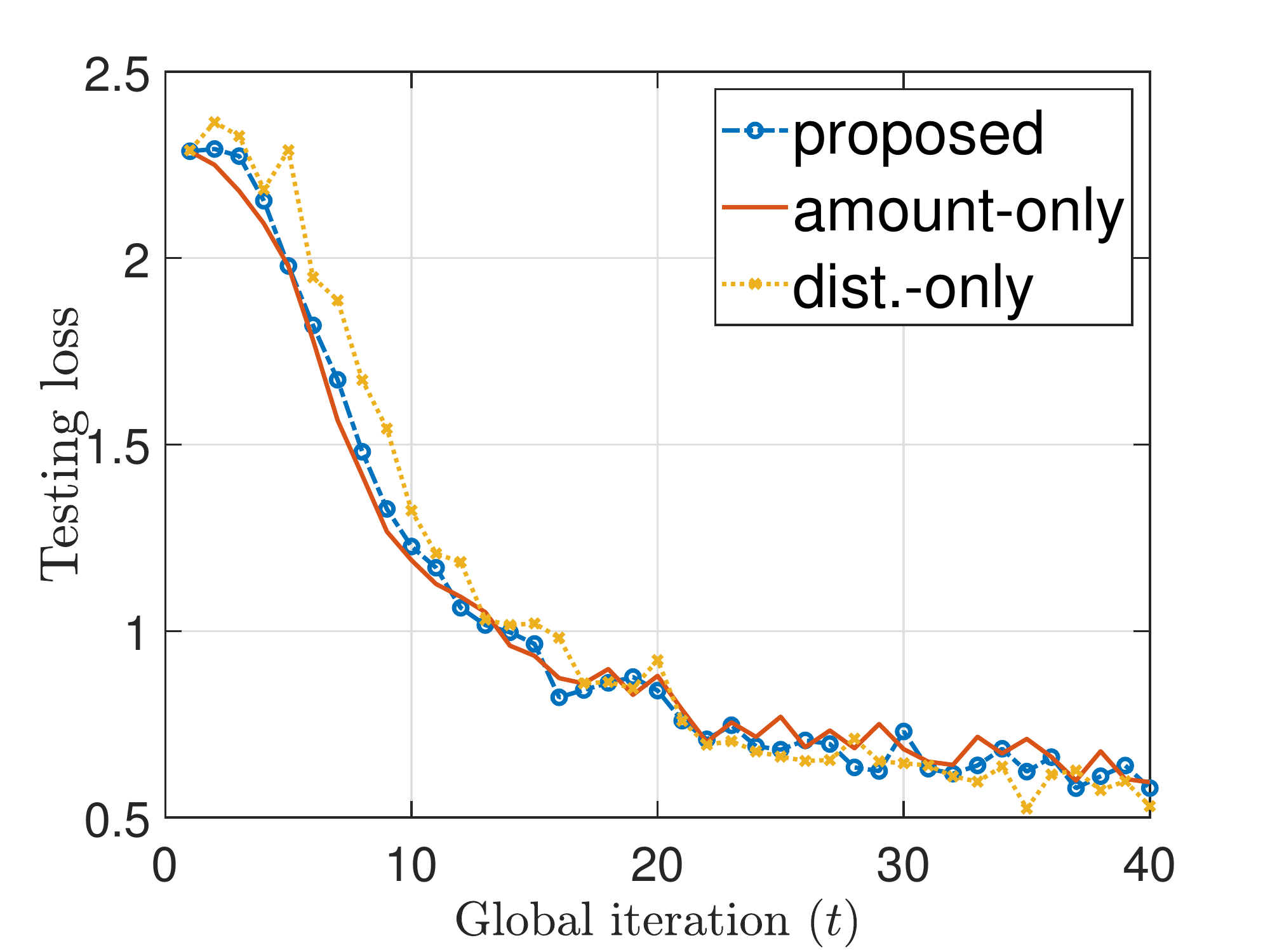}\label{fig:testLoss_dist0_niid}}
	\caption{Testing loss based on different data importance metrics. Most of the devices have i.i.d. data with digits 0 to 7 except $0.2K$ of them only contain data with digits $8$ and $9$.}
	\label{fig:metricCmpDiffDist}
	\hfill
\end{figure}

To validate the performance benefits of the proposed data importance metric $I_k(t)$, we compare \eqref{eq:iptc} with other metrics:
\begin{itemize}
	\item Amount-only metric, i.e., the first term in \eqref{eq:iptc}.
	\item distribution-only metric, i.e., the second term in \eqref{eq:iptc}.
\end{itemize}
Here, the feature vectors $\boldsymbol{x}$ and $\boldsymbol{y}_k$ are computed based on the digit-label distribution of the considered data.
The comparison of test accuracy is shown in Fig. \ref{fig:metricCmpDist8}, which confirms the design aspect of favoring those with higher number of newly arrived data.\footnote{In Figs \ref{fig:metricCmpDist8}, \ref{fig:metricCmpDiffDist}, the scaling factor $V$ is chosen to be sufficiently large to emphasize more on optimizing the time-average objective.}
A larger gap of test accuracy between the curves of amount-only and distribution-only methods can be observed at the early iterations, while the distribution-only method achieves a similar and even higher level of test accuracy than the others at the later iterations.
The reason behind the huge test accuracy difference is the highly unbalanced timing distribution that data generation at a device follows. In Fig. \ref{fig:metricCmpDiffDist}, we have shown the curves of testing loss under different timing distributions.
The testing loss difference between the two metrics becomes smaller in the scenario of uniformly distributed timings, since the amount of newly arrived samples per iteration is similar for all the devices. Moreover, the performance gain of distribution-only method at the later stage becomes more clear in Fig. \ref{fig:testLoss_dist0_niid} in the considered non-i.i.d. scenario. By combining the benefits of both metrics, the proposed data-importance metric is confirmed to be an effective measure to support an optimal scheduling design.
\section{Conclusions}
We investigated the problem of device scheduling in a FEEL system with random data generation at edge devices with energy and latency constraints. To deal with the system dynamics in data arrivals and energy consumption, we adopted Lyapunov optimization for designing a dynamic scheduling algorithm that maximizes the long-term data importance from scheduled device sets under constraints on energy consumption and per-round latency. 
The proposed method showed clear advantages in reducing energy consumption and achieving better learning performance as compared to baseline methods.

\bibliographystyle{IEEEtran}
\bibliography{ref.bib}

\end{document}